\documentclass[a4paper]{article}

\usepackage{INTERSPEECH2019}
\usepackage{multirow}
\usepackage{url}
\usepackage{xcolor}
\usepackage[normalem]{ulem}

\newcommand{\explabel}[1]{{\footnotesize\texttt{#1}}}

\title{An Unsupervised Autoregressive Model for Speech Representation Learning}
\name{Yu-An Chung, Wei-Ning Hsu, Hao Tang, James Glass}
\address{
  Computer Science and Artificial Intelligence Laboratory\\
  Massachusetts Institute of Technology\\
  Cambridge, MA 02139, USA}
\email{\{andyyuan,wnhsu,haotang,glass\}@mit.edu}

\begin{document}

\maketitle
 
\begin{abstract} 
This paper proposes a novel unsupervised autoregressive neural model for learning generic speech representations.
In contrast to other speech representation learning methods that aim to remove noise or speaker variabilities, ours is designed to preserve information for a wide range of downstream tasks.
In addition, the proposed model does not require any phonetic or word boundary labels, allowing the model to benefit from large quantities of unlabeled data.
Speech representations learned by our model significantly improve performance on both phone classification and speaker verification over the surface features and other supervised and unsupervised approaches.
Further analysis shows that different levels of speech information are captured by our model at different layers.
In particular, the lower layers tend to be more discriminative for speakers, while the upper layers provide more phonetic content.
\end{abstract}
\noindent\textbf{Index Terms}: speech representation learning, unsupervised learning

\section{Introduction}

Speech signals encompass a rich set of acoustic and linguistic properties, ranging from the individual lexical units, such as phonemes and words, to the characteristics of the speakers, their intent, or even their mental status.
However, these high-level properties of speech are poorly captured by the surface features, such as the amplitudes of a wave signal, log Mel spectrograms, or Mel frequency cepstral coefficients.
The goal of speech representation learning is to find a transformation from the surface features that makes high-level properties of speech more accessible to downstream tasks.

In this work, we propose an autoregressive model for learning speech representations that can be transferred to different tasks across different datasets.
Our model is able to retain much information from the surface features, allowing a wide range of tasks across different datasets to benefit from the learned representations, while also being unsupervised and able to leverage large quantities of unlabeled data.
As a first step, we focus on learning general speech representations from log Mel spectrograms, but it is straightforward to extend our approach to the amplitudes of the wave signals.

We use linear separability~(or separability with a shallow network) to define the accessibility of information for the downstream tasks.
Others~\cite{schatz2013evaluating} have argued that there are many nuisance factors that might affect the performance of linear classifiers, and have proposed to use a contrastive loss for evaluation.
However, there has been evidence~\cite{settle2016discriminative} and theories~\cite{arora2019theoretical} supporting the idea that low contrastive loss implies the existence of a linear classifier with low error.
In other words, we aim to learn speech representations that allow linear classifiers to perform well on many downstream tasks.

When the downstream tasks are known, supervised learning, specifically multitask learning~\cite{caruana1997multitask}, is the most successful approach for learning specialized representations of those particular tasks.
In general, however, when a transformation is trained against a certain set of tasks, information independent of the tasks~(such as noise or speaker variability, depending on the tasks) tends to be discarded after training~\cite{tishby1999information}.
We risk discarding useful information for other unseen tasks when learning representations in a supervised fashion.
Instead of having targeted tasks in advance, we focus on learning representations for a wide range of, potentially unknown, tasks.
Due to the required generality, it is necessary to retain in the representations as much information about the original signal as possible.
Two commonly used loss functions, i.e., the autoencoding and autoregressive loss functions, satisfy this criterion.
However, when no additional constraints are imposed, there is a trivial solution, the identity mapping, for the autoencoding loss function.
This makes the autoregressive loss more appealing, because no additional techniques, such as denoising~\cite{vincent2010stacked}, are required to avoid the trivial solution as for the autoencoding loss.
The autoregressive approach also does not require other types of linguistic constraints, such as phonetic or word boundaries~\cite{kamper2016deep}.

The autoregressive loss belongs to a large family of self-supervised loss functions~\cite{wang2015unsupervised,doersch2015unsupervised,larsson2017colorization}.
There also exists some work on unsupervised speech representation learning~\cite{chorowski2019unsupervised,chung2018speech2vec,milde2018unspeech,hsu2017unsupervised,hsu2017learning,chung2016audio}.
However, none of the studies are able to show the transferability of the learned representations across different datasets.
Our work is largely motivated by the recent success in transfer learning from large-scale pre-trained language models~\cite{peters2018deep,howard2018universal,radford2018improving,devlin2018bert}, and we aim to learn general speech representations that can be transferred to different tasks across different datasets.

\section{Models}

We propose a novel autoregressive architecture, which we call Autoregressive Predictive Coding~(APC), for unsupervised speech representation learning.
Predictive coding on wave samples~\cite{schroeder1985code} has a long and influential history in speech processing, and its recent neural version~\cite{oord2016wavenet} and variants, such as Contrastive Predictive Coding~(CPC)~\cite{oord2018representation}, have also been used to learn speech representation~\cite{chorowski2019unsupervised}.
In contrast to these studies, our work mainly focus on predicting the spectrum of a future frame rather than a wave sample.
We will briefly review CPC here and compare extensively with it in Section~\ref{sec:exps}.

\subsection{Autoregressive Predictive Coding}

The methodology of the proposed APC model is largely inspired by language models~(LMs) for text, which are typically a probability distribution over sequences of~$N$ tokens~$(t_{1}, t_{2}, ..., t_{N})$.
Given such a sequence, an LM assigns a probability~$P(t_{1}, t_{2}, ..., t_{N})$ to the whole sequence by modeling the probability of token~$t_{k}$ given the history~$(t_{1}, t_{2}, ..., t_{k - 1})$:
\begin{equation}
  P(t_{1}, t_{2}, ..., t_{N}) = \prod_{k = 1}^{N}P(t_{k}\mid t_{1}, t_{2}, ..., t_{k - 1}).
\end{equation}
It is trained by minimizing the negative log-likelihood:
\begin{equation}
  \sum_{k = 1}^{N}-\log P(t_{1}, t_{2}, ..., t_{k - 1}; \theta_{t}, \theta_{\mathrm{rnn}}, \theta_{s}),
\end{equation}
where the parameters to be optimized are~$\theta_{t}$,~$\theta_{\mathrm{rnn}}$, and~$\theta_{s}$.
$\theta_{t}$ is a look-up table that maps each token into a vector of fixed dimensionality.
$\theta_{\mathrm{rnn}}$ is a Recurrent Neural Network~(RNN) used to summarize the sequence history up to the current time step.
$\theta_{s}$ is a Softmax layer appended at the output of each RNN time step for estimating probability distribution over the tokens.
Language modeling is a general task that requires the understanding of many aspects in language in order to perform well.

Following most of the neural LMs in the literature, we use an RNN~\cite{mikolov2010recurrent} for modeling the temporal information within an acoustic sequence.
For speech data, each token~$t_{k}$ corresponds to a frame rather than a word or character token, hence we do not need the look-up table~$\theta_{t}$ as we do in LMs and directly feed each frame into the RNN~$\theta_{\mathrm{rnn}}$.
Since there does not exist a finite set of target tokens~(such as the vocabulary set as in text), we choose to replace the Softmax layer with a regression layer~$\theta_{r}$.
In other words, the RNN output at each time step attempts to directly fit the target frame with a linear mapping.
The learnable parameters in APCs are~$\theta_{\mathrm{rnn}}$ and~$\theta_{r}$.

Given the history~$(t_{1}, t_{2}, ..., t_{k - 1})$, an LM aims to maximize the probability of the next token to be the~$t_{k}$ in the data.
However, for APCs, exploiting the local smoothness of the speech signal might be sufficient to predict the next frame.
To encourage APCs to infer more global structures rather than the local information in the signals, we ask the model to predict a frame~$n$ steps ahead of the current one.
In other words, given an utterance represented as a sequence of acoustic feature vectors~$(x_{1}, x_{2}, ..., x_{T})$, the RNN processes each sequence element~$x_{t}$ one at a time and outputs a prediction~$y_{t}$, where~$x_{t}$ and~$y_{t}$ have the same dimensionality.
The model
is optimized by minimizing the L1 loss~(as is done when predicting spectral frames in some speech synthesis models~\cite{wang2017tacotron,chung2019semi}.) between the input sequence~$(x_{1}, x_{2}, ..., x_{T})$ and the predicted sequence~$(y_{1}, y_{2}, ..., y_{T})$:
\begin{equation}
  \sum_{i = 1}^{T - n}\lvert x_{i + n} - y_{i}\rvert.
\end{equation}

\subsection{Contrastive Predictive Coding}
Instead of learning to predict future frames like APCs, Contrastive Predictive Coding~(CPC)~\cite{oord2018representation} aims to learn representations that separates the target future frame~$x_{i + n}$ and randomly sampled negative frames~$\{\tilde{x}\}$, given a context~$h_i = (x_1, x_2, ..., x_i)$.

Specifically, CPC consists of three modules: a frame encoder~$E_{frm}$, a uni-directional RNN~$E_{ctx}$, and a scoring function~$f$.
A sequence of frames is first encoded to a sequence of frame representations~$z_i = E_{frm}(x_i)$ using the frame encoder.
The encoded sequence is then passed to the recurrent context encoder to obtain a sequence of context representations~$(c_1, c_2, ..., c_T)$, where~$c_i$ is a fixed-dimensional representation computed from~$E_{ctx}(z_1, z_2, ..., z_i)$.
The scoring function assigns a positive scalar to a pair of frame and context, formulated as~$f(x, h) = \exp \left(z^T W c \right)$, where~$z$ is the frame representation of~$x$, and~$c$ is the context representation of~$h$.

Suppose the target frame is~$n$ steps away.
Given a context~$h_i$, the target future frame~$x_{i + n}$, and a collection of negative frames~$\tilde{X}$, CPC jointly optimizes the three modules by minimizing a contrastive loss:
\begin{equation}
  \mathcal{L}_n(h_i, x_{i + n}, \tilde{X}) = \log \dfrac{f(x_{i + n}, h_i)}
  {\sum_{x \in \tilde{X} \cup \{x_{i + n}\}\ } f(x, h_i)}.
\end{equation}
As shown in~\cite{oord2018representation}, minimizing this loss will result in~$f(x, h)$ estimating the density ratio~$p_n(x \mid h) / q(x)$, where~$p_n$ denotes the conditional distribution of~$x$ at~$n$ steps ahead of the given context~$h$, and~$q$ is the proposal distribution where negative samples are drawn from.
In other words, the choice of the number of steps ahead and the proposal distribution would both affect the estimated target density ratio, and therefore would change what is learned in the representations~$z$ and~$c$.
For example, using a proposal distribution that draws samples from the same sequence as the target frame would encourage the model to learn the phonetic content but not the speaker information, because the latter do not help distinguishing a target frame from negative ones.
We will study such differences in our experiments.

Both CPC and the proposed APCs consider the sequential structures of speech, and predict information about future frames.
However, the two models differ significantly in the type of information the corresponding loss function enforces them to capture.
While CPC representations are encouraged to focus on information that is most discriminative between the target and negative frames, APCs have to encode information sufficient for predicting the target frame, and are allowed to only discard information that is common across the train dataset.

\section{Experiments}
\label{sec:exps}

In this section, we empirically demonstrate the effectiveness of the learned representations from the proposed APC model.
Since phone and speaker information are two of the most important characteristics that differentiate one speech utterance from another, we choose to use phone classification and speaker verification to examine how much phone and speaker information are captured by the representations.

\subsection{Datasets}

We use the LibriSpeech corpus~\cite{panayotov2015librispeech} for training the feature extractors~(all APC and CPC models).
Specifically, the~360-hour subset, which contains~921 speakers in total, is used.
We use 80-dimensional log Mel spectrograms~(normalized to zero mean and unit variance per speaker) as input features.

An ideal feature extractor should extract representations that generalize to datasets of different domains.
To examine the robustness to shift in domains, rather than on the LibriSpeech test set, we conduct phone classification and speaker verification on the Wall Street Journal~(WSJ)~\cite{paul1992design} and TIMIT corpora.
For phone classification, we follow the standard split of WSJ, use 90\% of \texttt{si284} for training, use the rest of the 10\% for development, and report numbers on \texttt{dev93}.
The phone alignments are generated with a speaker adapted GMM-HMM model.
For speaker verification, we follow the standard split of TIMIT, use the training set for training the universal background model, the i-vector extractor~\cite{dehak2011front}, a linear discriminant analysis~(LDA) model.
We follow the standard practice of speaker verification and only consider female-female and male-male pairs in the 50-speaker development set.
We note that speaker verification on TIMIT is not common, and we mainly use it to check if the representations contain speaker information.

\subsection{Implementations}

We model our APCs with a multi-layer unidirectional LSTM~\cite{hochreiter1997long} network with residual connections~\cite{he2016deep}
between two consecutive layers as is done in~\cite{wu2016google}, and the dimensionality of each layer is~512.
For CPC, we follow the implementation for the context encoder and the scoring function in~\cite{oord2018representation}, but change the acoustic feature~$x$ from a window of~400 samples~(25ms) to a~80-dimensional vector of Mel spectra computed from that segment, and replace the~5-layer strided Convolutional Neural Network with a~3-layer,~512-dim fully-connected neural network with ReLU activations for the frame encoder.
Such modification aims for a fairer comparison between APC and CPC models in terms of their training objectives, while eliminating the source of variation due to the choice of acoustic features.
All APC and CPC models~(except \explabel{cpc-ctx-exhaust}, which we will describe more below) are trained for~100 epochs using the Adam optimizer~\cite{kingma2015adam} with a batch size of~32 and an initial learning rate of~$10^{-3}$.

Note that the proposed approach is unsupervised, and we do not and should not tune hyperparameters according to the downstream tasks.
The goal of hyperparameter tuning is to show how the hyperparameters affect what is learned in the speech representations.
Recall that we define the accessibility of categorical information as the linear separability among classes.
For phone classification, we simply use a linear classifier to predict the phoneme classes for each frame.
The frame error rates indicate how much phonetic content is contained in the speech representations.
Similarly, for speaker verification, we train an LDA model on top of the speech representations.

\subsection{Phone Classification}

Table~\ref{tab:phone_cpc-apc} compares APCs with a series of CPC models that use different training variants.
Phone error rates~(PER) are reported, and each of the first four rows corresponds to a CPC variant.
We use \explabel{cpc-n9all} to denote a CPC model that draws~9 negative samples from utterances within the same minibatch, and \explabel{cpc-n9same} to denote a CPC model that draws~9 negative samples from the same utterance.
For both \explabel{cpc-n9all} and \explabel{cpc-n9same}, we take the outputs of the frame encoder~(i.e., the outputs of the 3-layer fully-connected neural networks) as the extracted features and feed them to the linear classifier.
The training approach of \explabel{cpc-ctx-n9same} is the same as \explabel{cpc-n9same}, except that the RNN outputs are taken as the extracted features instead of the frame encoder outputs.
We use \explabel{ctx}, short for context, to indicate such difference.
The final CPC variant we try is \explabel{cpc-ctx-exhaust}, which follows the exact same training procedure in~\cite{oord2018representation} that combines contrastive losses for all steps~$k \leq n$ with equal weights for training~(i.e., $\sum_{k=1}^n \mathcal{L}_n$), uses all non-target samples in a minibatch as negative samples, and are trained with mini-batches of~8 utterances that are randomly chuncked to 128 frames each.
For APCs, the outputs of the last RNN layer are taken as the extracted features.
All models in Table~\ref{tab:phone_cpc-apc} consist of one RNN layer, and the effect of predicting different time steps ahead is also investigated.
\begin{table}[htbp]
  \centering
  \caption{
    Comparing APCs with a series of CPC models on phone classification.
    PERs are reported.}
  \label{tab:phone_cpc-apc}
  \begin{tabular}{lcccc}
    \toprule
    \multirow{2}{*}{Method}  &  \multicolumn{4}{c}{\#(step)}  \\
    \cmidrule(lr){2-5}
                         &   2    &   5    &   10   &   20    \\
    \midrule
    \midrule
    \explabel{cpc-n9all}
                         & 51.3   &  48.8  &  50.8  &  54.6   \\
    \explabel{cpc-n9same}
                         & 47.5   &  48.2  &  50.0  &  53.0   \\
    \explabel{cpc-ctx-n9same}
                         & 42.1   &  46.1  &  48.8  &  53.8   \\
    \explabel{cpc-ctx-exhaust}
                         & 42.9   &  43.1  &  45.6  &  49.1   \\
    apc~(proposed)       & 36.5   &  35.6  &  35.4  &  37.7   \\
    \bottomrule
  \end{tabular}
\end{table}

\newcommand{\myparagraph}[1]{\vspace{.4em} \noindent \textbf{#1}\ }
\myparagraph{A comparison of models from the CPC-family.}
From Table~\ref{tab:phone_cpc-apc}, we observe that \explabel{cpc-n9same} outperforms \explabel{cpc-n9all} across all time steps we try.
This is an expected outcome, since for \explabel{cpc-n9all}, the negative samples are drawn from different utterances within a minibatch that could possibly be uttered by different speakers, and thus \explabel{cpc-n9all} is not required to really capture phonetic content to differentiate the positive and negative samples.
In contrast, \explabel{cpc-n9same} draws negative samples from the same utterance, and in such case, speaker information is identical for each sample and \explabel{cpc-n9same} is forced to learn other non-trivial features such as phone information so as to differentiate positive and negative samples.
In addition, we find that representations extracted from RNN contain more phonetic content than those extracted from the frame encoder, as \explabel{cpc-ctx-n9same} often outperforms \explabel{cpc-n9same} especially when the number of steps to the target is small.
By using all non-target samples as negative samples from the minibatch, \explabel{cpc-ctx-exhaust} further lowers the PER, suggesting that richer phonetic content is learned in the representations.

\myparagraph{Comparing CPC with APC.}
Our APCs, as shown in the last row in Table~\ref{tab:phone_cpc-apc}, significantly outperform all CPC models in spite of its much simpler architecture and training approach.
These results demonstrate that more phonetic content is immediately accessible from a linear classifier in the representations extracted by APCs compared to CPC models.

There are other aspects of APCs worth investigating.
In Table~\ref{tab:phone_apc-logMel}, we present the phone classification results of using deeper RNNs for APCs and with more target time steps.
For all APC models, we take the outputs of the last RNN layer as the extracted features.
Three supervised baselines, a linear classifier, a~1-layer multi-layer perceptron~(MLP), and a~3-layer MLP, are implemented, taking the \textit{surface features}, i.e., spectrograms, as input features.
For MLPs, each layer consist of~512 units with ReLU activations.
These three baselines are meant to help us understand how accessible the phonetic content is from the surface features, even under some amount of nonlinear transformations.
We also include the best number of CPC models from Table~\ref{tab:phone_cpc-apc} to bridge the two tables.
\begin{table}[htbp]
  \centering
  \caption{
    PERs on phone classification.
    All features are fed to a linear classifier unless otherwise stated.
    The number of steps to the target \#(steps) is not relevant in the first four rows.}
  \label{tab:phone_apc-logMel}
  \begin{tabular}{lcccccc}
    \toprule
    \multirow{2}{*}{Method}         &  \multicolumn{6}{c}{\#(step)}  \\
    \cmidrule(lr){2-7}
        &   1    &   2    &   3    &   5    &   10   &   20          \\
    \midrule
    \midrule
    Mel          &  \multicolumn{6}{c}{50.0}    \\
    Mel + MLP-1  &  \multicolumn{6}{c}{43.4}    \\
    Mel + MLP-3  &  \multicolumn{6}{c}{41.3}    \\
    cpc best     &  \multicolumn{6}{c}{42.1}    \\
    apc 1-layer   &  39.4  &  36.5  &  35.4  &  35.6  &  35.4  &  37.7  \\
    apc 2-layer   &  38.5  &  34.6  &  35.9  &  35.7  &  34.6  &  38.8  \\
    apc 3-layer   &  37.2  &  36.7  &  33.5  &  36.1  &  37.1  &  38.8  \\
    apc 4-layer   &  36.2  &  34.4  &  34.5  &  35.3  &  36.9  &  39.6  \\
    \bottomrule
  \end{tabular}
\end{table}

\myparagraph{Surface features with non-linear phone classifier.}
From Table~\ref{tab:phone_apc-logMel}, we observe that incorporating non-linearity in the phone classifier does improve PER\footnote{The best performing supervised~3-layer LSTM with minimal lookahead on this particular task can achieve~16.3~\cite{tang2018training}.}.
When using a~3-layer MLP as the classifier, the surface features are transformed into higher-level representations that are more linearly separable than the best CPC features.
However, we can see there is still a significant gap between the transformed spectrogram representations with features extracted by APC models.

\myparagraph{A comparison of APC models.}
Overall, we find that deeper APC models produce better representations especially for small \#(steps).
There also exists a sweet spot when we vary the amount of time steps to the target for APC models to predict---the PER continues to drop as we increase \#(steps) until a certain point, which is usually when \#(steps) equals~3; after that the PER begins to increase as \#(steps) increases.

\subsection{Speaker Verification}

For speaker verification, we compare APCs with the i-vector representation.
We train a GMM with~256 components as the universal background model on the TIMIT training set.
We then extract~100-dimensional i-vectors and project them down to~24 dimensions with LDA trained on the training set.
The cosine similarity is used for evaluation.
We also include the best results from all CPC models.
The equal error rates~(EER) on speaker verification are presented in Table~\ref{tab:spk_apc-logMel}.
Same as what we do in the phone classification experiments, the outputs of the last RNN layer are taken as the extracted representations.
The representation of the entire utterance is a simple average of the frame representations.
For the last two rows , i.e., apc~3-layer-1 and apc~3-layer-2, it means that we take the outputs of the first and the second RNN layer as the extracted representations.
We explain our motivation of doing so below.
\begin{table}[htbp]
  \centering
  \caption{
    EER on speaker verification.
    The number of steps to the target \#(steps) is not relevant for the first two rows.}
  \label{tab:spk_apc-logMel}
  \begin{tabular}{lcccccc}
    \toprule
    \multirow{2}{*}{Method}         &  \multicolumn{6}{c}{\#(step)}     \\
    \cmidrule(lr){2-7}
                  &   1    &   2    &   3    &   5    &   10   &   20   \\
    \midrule
    \midrule
    i-vector               &  \multicolumn{6}{c}{6.64}    \\
    cpc best               &  \multicolumn{6}{c}{5.00}    \\
    apc 1-layer   &  4.71  &  4.07  &  4.14  &  4.14  &  5.14  &  5.29  \\
    apc 2-layer   &  4.71  &  4.64  &  5.71  &  4.86  &  5.57  &  6.07  \\
    apc 3-layer   &  5.21  &  4.93  &  4.43  &  4.57  &  5.79  &  6.21  \\
    apc 3-layer-1 &  3.43  &  3.86  &  3.79  &  3.86  &  4.07  &  4.86  \\
    apc 3-layer-2 &  3.79  &  4.64  &  4.14  &  4.29  &  5.14  &  5.00  \\
    \bottomrule
  \end{tabular}
\end{table}

\myparagraph{Comparing APC with i-vector and CPC.}
From Table~\ref{tab:spk_apc-logMel}, we can see that the best CPC model outperforms the i-vector baseline, and APCs further outperform CPC when \#(steps) is smaller than~10.
This demonstrates that representations learned by APCs contain not only phonetic information but also speaker information.

\myparagraph{Speaker information across different APC layers.}
Unlike phone classification, where we find increasing the depth of APCs improve PER, deeper APCs somehow performs worse in speaker verification.
Studies have shown that in a deep LM, lower layers tend to focus more on local syntax, while the upper layers usually induce more semantic content~\cite{peters2018dissecting}.
Motivated by the fact that LMs for text could exhibit different kinds of information across different layers, we are interested in investigating whether other layers besides the last one contain more information of our interest, that is, the speaker information.
Specifically, instead of taking the outputs of the last RNN layer of apc~3-layer, we try using the outputs of the first and second RNN layers of it to perform speaker verification, denoted by apc~3-layer-1 and apc~3-layer-2 in Table~\ref{tab:spk_apc-logMel}, respectively.
Surprisingly, for all \#(steps), we see that apc~3-layer-1 consistently outperforms apc~3-layer-2, which further outperforms apc~3-layer.
This indicates that lower layers indeed contain more speaker information than higher layers, or at least the speaker information is represented in a more accessible form in lower layers.
Additionally, we observe that apc~3-layer-1 outperforms apc~1-layer and apc~3-layer-2 outperforms apc~2-layer although the representations are extracted from the same RNN depth.
Combining all of our observations from both tasks, we conclude that a deep APC is a very powerful speech feature extractor, whose higher layers capture phonetic information while more speaker information resides in its lower layers.


\section{Discussions}

We propose Autoregressive Predictive Coding~(APC) for unsupervised speech representation learning.
The backbone of APC is a deep LSTM network, and the model is trained in an autoregressive fashion.
We introduce a time shifting factor that asks the model to predict further steps ahead of the current frame during training in order to encourage it to discover more general structures rather than the local ones within the speech signal.
Our experimental results show that the number of steps to the target frame controls what is learned in the representation.
How this hyperparameter is set depends on how the representation is going to be used and can be thought of as a prior.

Transfer learning from large-scale pre-trained LMs has shown great success recently, and we believe it is promising and useful to develop similar transfer learning techniques for the domain of speech and audio.
APC proposed in this work is our first step towards this goal.
Despite its simplicity, APCs have demonstrated a strong capability of extracting useful phone and speaker information through our experiments.
In the future, we are interested in training APCs on larger and probably noisier corpora and testing the extracted features on other speech-related tasks.
Furthermore, in this work we only take outputs from a specific layer from APC models as input features for a downstream task.
However, as indicated in our experimental results that different layers may focus on capturing different aspects of speech information~(e.g., lower layers are shown to contain richer speaker information than the upper layers), it is potentially beneficial to combine all internal representations across different layers and simultaneously expose all of them to a downstream model.
This allows the model to select which the combination~(e.g., through a set of learnable weights as in done in ELMo~\cite{peters2018deep}) of all representations most useful for an end task.
From the point of view of model interpretability, it is also important to analyze how the internal representations in a deep APC are transformed across layers from capturing speaker information to capturing phonetic information.

\bibliographystyle{IEEEtran}
\bibliography{mybib}

\end{document}